# A Computational Model for Situated Task Learning with Interactive Instruction

Shiwali Mohan (shiwali@umich.edu), James Kirk (jrkirk@umich.edu), John Laird (laird@umich.edu)
Computer Science and Engineering, University of Michigan
2260 Hayward Street, Ann Arbor, MI 48109 USA

**Abstract**

Learning novel tasks is a complex cognitive activity requiring the learner to acquire diverse declarative and procedural knowledge. Prior ACT-R models of acquiring task knowledge from instruction focused on learning procedural knowledge from declarative instructions encoded in semantic memory. In this paper, we identify the requirements for designing computational models that learn task knowledge from situated task-oriented interactions with an expert and then describe and evaluate a model of learning from situated interactive instruction that is implemented in the Soar cognitive architecture.

**Keywords:** task learning, instruction, cognitive architecture.

## Introduction

Over the last 30 years, the field has made great progress in modeling human behavior, not only in developing computational models of specific tasks, but also in developing the underlying cognitive architecture that is a theory of the fixed memories, representations, and processing units shared across those tasks. However, we have only limited understanding as to how people acquire the knowledge required to perform those tasks, and invariably we rely on humans to encode the necessary procedural knowledge. This leaves unanswered why a human would have the specific procedural knowledge that was encoded in the model. Additional sources of information, such as brain activity data and assumptions about rationality and optimality provide constraints, but they leave unanswered what knowledge and computational processes humans use for interactively acquiring, assimilating, and using knowledge for novel tasks.

There has been only limited research on using recipe-style task instructions, written in constrained forms of English (Anderson, 2007). In these approaches, the model stores the instructions in long-term declarative memory, and then during task performance, the model interprets the instructions, and through execution, converts them to procedural knowledge that is available for future task performance. These models support only one-time and one-way interactions from instructor to student using fixed vocabularies and are limited to internal reasoning tasks.

In our approach, *situated interactive instruction*, the learner model (embodied in a table-top robot) and the expert human are simultaneously embedded in a shared, real-world domain. Through mixed-initiative, natural language instructions the learner acquires various concepts and behaviors useful in the domain. In our previous work (Mohan et al. 2012), we studied interactive acquisition of *basic* concepts such as attributes of objects (color: *red*, size: *large*), spatial relationships (*right-of*), and simple actions (*move*). The learner also acquires new vocabulary of adjectives, nouns, prepositions, and verbs that are grounded in basic concepts and can be used in interactions. It is implemented in Soar (Laird, 2012) and is pre-encoded with only limited initial knowledge of the domain. Thus, it must learn not only procedural knowledge (actions and control knowledge), but also object descriptions and spatial relations.

In this paper, we demonstrate that our learning mechanisms are sufficient for learning *complex* concepts such as hierarchical task knowledge that are novel compositions of the basic concepts it has already learned. The procedural and declarative knowledge acquired for novel tasks can be extended for solving simple puzzles and playing games. Although our research is not yet at the point where we can quantitatively compare our model with human task learning, this paper makes the following contributions to the cognitive modeling community:

1. It identifies the computational and behavioral requirements for learning from situated interactive instruction.
2. It describes an implemented model of learning from situated interactive instruction, realized within the constraints of a cognitive architecture. Our model presents a theory of how linguistic, declarative, and procedural task knowledge are represented in semantic, episodic, and procedural memories, and how these types of knowledge are learned.

## Situated Interactive Instruction

Mixed-initiative, task-oriented, natural language, interactions arise naturally in situations where an expert guides a novice to perform a novel task. The facilitator expert and the primary learner form a system of joint learning, which distributes the onus of learning between both participants. The expert takes initiative in identifying the relevant objects and relationships in the shared environment and structuring and decomposing tasks. The learner takes initiative in actively interpreting the instructions, applying them to the current situation, analyzing successes and failures, and posing relevant queries that elicit useful information from the expert. To model a learner that interactively acquires knowledge, several complex aspects of cognition (described below) have to be addressed.

### Requirements

The learner is embedded in an environment and must maintain an ongoing interaction with the expert. Along with the basic perceive-decide-act cycle, in which the learner perceives objects and their relevant properties and relation-ships, decides its next goal, and manipulates the environment in accordance with the goal, it must also encode the following computational mechanisms.

**R1. Integrative Interaction.** Tutorial interactions are highly contextual. A complex interaction unfolds as participants negotiate the meaning of utterances, accumulate common ground, and act in the environment. To maintain the state of the ongoing interaction, the learner must employ a *task-oriented* interaction model. It should allow both participants to change the focus of interaction to different aspects of the task based on their goals.

**R2. Referential Comprehension.** To comprehend utterances, the learner must transform linguistic symbols in the expert's utterances to its internal representation of the environmental state and knowledge of the domain. Designing computational models for such comprehension poses a significant challenge because utterances can be linguistically ambiguous, requiring the learner to exploit extra-linguistic context for comprehension. Our previous work on learning how to associate linguistic symbols in novel concepts acquired through instruction (Mohan et al. 2012) and generating grounded interpretations of utterances using word-concept associations (Mohan and Laird 2012) addresses some issues related to referential comprehension.

**R3. Situated Learning.** The experience of interactive execution of novel tasks is rich in information about the relevant aspects of the task. The learner should extract diverse task knowledge – linguistic, perceptual, spatial, semantic, and procedural by analyzing the experience of task execution and interactions.

**R4. Active, incremental learning.** Interactive learning affords an important advantage. An intelligent learner contributes to its own learning by asking questions that aid its understanding of the task. Replies from the expert are integrated with the learner's prior knowledge of the task. To design a learner that demonstrates such behavior, three questions have to be answered: *when should a question be asked?*; *what question should be asked?*; and *how is the reply integrated with prior knowledge?* We show that meta-cognitive analysis performed during impasse resolution in Soar can inform all of these questions.

## Desirable Behaviors

Learning from interactive instruction is a complex cognitive activity. There is a wide range of behaviors that are expected from a competent learner. In this paper, we focus on the following desirable behaviors.

**B1. General knowledge from specific experience** Information in tutorial instructions usually pertains to the current situation. The participants communicate about the current state of the task within the context of currently perceptible objects and their state and relationships and the actions that can be taken in the current state of the environment. The learner should be able to learn general domain knowledge from few highly specific examples in tutorial interactions. Analytical learning methods such as explanation-based learning (EBL; Mitchell et al. 1986) are useful for deriving general procedural knowledge for task execution from few specific examples of behavior.

**B2. Flexible instruction.** The learning mechanisms we describe do not impose a strict order on how the learner is taught new concepts. This gives flexibility to the expert to structure the nature of instruction. An expert can teach basic concepts before teaching complex concepts that require knowledge of the basic concepts. However, the expert may not know or remember the state of the learner's knowledge. In situations when the learner is learning complex concepts but lacks required knowledge of basic concepts, it will take the initiative and guide the interaction to acquire the relevant basic concepts first. We show that our mechanisms are sufficient for learning task knowledge starting from varying prior knowledge states.

**B3. Extendible behavior.** The task knowledge acquired by the learner can be extended to new tasks by providing instructions about the additional constraints. After acquiring general procedural knowledge of *move*, the learner can be taught to solve puzzles like Tower of Hanoi and play games like Tic-Tac-Toe by giving instructions about the relevant parameters and constraints of legal actions and goals.

## Domain

The learner is embodied in a table-top robot (Figure 1) that can perceive (via a Kinect) and manipulate small blocks. The workspace contains four locations: *pantry*, *garbage*, *table*, and *stove*. These locations have associated simulated functionalities. The learner can perform several primitive actions in the domain including `open/close (pantry/stove)`, `turn-on/turn-off (stove)`, `pick-up/put-down (object)`.

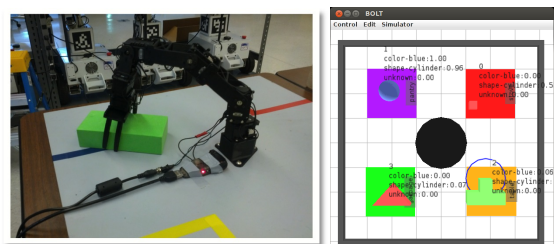

Figure 1: The table-top robot assembly and the simulator.

## Preliminaries

We begin with an overview of Soar and its mechanisms relevant to our implementation and brief descriptions of the models of referential comprehension and integrative interaction. Later, we describe our task acquisition model in detail.

### Soar Cognitive Architecture

*Working memory* encodes beliefs derived from the perceptions of the environment and the learner's experiential knowledge encoded in its long-term memories. The beliefs derived from perceptions consist of all objects currently in the visual field. The *spatial visual system* (SVS) allows the learner to extract binary spatial predicates that describe the alignment of objects along the three axes and the distances between objects. Spatial relationships, such as *right-of,* are learned as a composition of spatial predicates.

*Procedural memory* holds long-term knowledge for generating and controlling behavior, such as interaction management, language comprehension, and action and task execution knowledge. This knowledge is encoded as if-then rules that propose, select, and apply *operators*, which are the locus of decision-making. If the knowledge for selecting or applying an operator is incomplete, an *impasse* arises and a substate is generated, in which operators can be selected to resolve the impasse through methods such as task-decomposition and planning. *Chunking* creates new rules that summarize the processing in a substate.

*Semantic memory* stores context-independent declarative concepts about the environment, which are represented as sets of relational graph structures. Concepts are retrieved by creating a *cue* in a working memory buffer. The best match (biased by recency and frequency) is then added to working memory. *Episodic memory* stores the learner's experiences of task execution and interactions. It enables the learner to recall the context and temporal relations of past experiences using cue-based retrievals. The best match (biased by recency) is retrieved and added to working memory.

**Integrative Interaction Model**

The learner uses an interaction model (Mohan et al. 2012) based on the theory of task-oriented discourse (requirement R1) by Grosz (1986). It organizes the expert-learner utterances in discourse as hierarchical *segments* with *purposes* aligned with the goals of the task. The state of discourse is represented as a stack of segments. This model allows both participants to change the focus of the interaction by introducing a new segment with a specific purpose that will be useful in achieving the goal of the participant.

**Referential Comprehension Model**

To gain useful information from an utterance, the learner must ground linguistic references to symbols arising from perceptions, spatial reasoning, and task knowledge (requirement R2). We use the term *map* for structures in semantic memory that encode how linguistic symbols (nouns/adjectives, prepositions, and verbs) are associated with referent concepts - perceptual symbols, spatial compositions, and task concept networks. To ground a sentence, the indexing process (Mohan & Laird, 2012) retrieves relevant maps from semantic memory so that it can connect the linguistic terms with their referents. If the terms are successfully mapped, the learner uses constraints derived from the referents, the current environmental state, domain models, and the interaction context to create a grounded representation of the utterance.

**Task Knowledge Representation**

A task is defined by a goal state that requires primitive actions by the learner in the environment. In our formulation tasks are referred to using verbs. For example, *"move the blue cylinder to the pantry"* is a task where the learner must execute primitive actions *pick up the blue cylinder* and *put the cylinder in the pantry* to establish the spatial-relationship *in(blue cylinder, pantry)*. A task (and thus a verb) may be composed of other constituent tasks (and verbs) that ultimately ground out in primitive actions. To *store* an object, the learner can *open the pantry* and execute the constituent task *move* such that it establishes the goal of *store*. In learning the hierarchical knowledge for a task, the learner must acquire the following concepts.

A **goal** is a composition of predicates that encode the state of objects, spatial relationships between objects, etc. that determine if the task has been successfully executed. The **problem-space** is a set of operators that are sufficient for successfully completing a task. This set can contain primitive actions (*open*, *pick-up, put-down*) or constituent tasks (*move*). The goal and problem-space are learned through instruction and encoded declaratively in the learner's semantic memory as a **task concept network (tCN)**. An example tCN (nodes M22, G22, PR1, O2 PR2, PS22, A12) for *store* is in Figure 2.

A **map** connects the syntactical structure of the verb that describes the action and its argument structure to the tCN. This knowledge is useful for referential comprehension of verbs while generating interpretations of action commands. An example is in Figure 2 (nodes L22, O1) connects to the tCN via the map node M22.

**Behavior** is produced by rules that select the correct operator to execute given the current environmental context and goals. A behavior is abstracted as a task operator (nodes P22, O1 in Figure 2). The tCN constrains how the task operator *op1* is instantiated given the contents of the action command and goals. For example, *argument1* of *op1* is constrained to be the object that satisfies the description in the direct-object argument of the verb *store*.

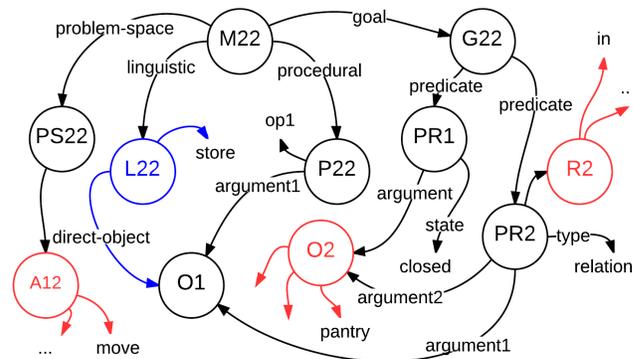

Figure 2: Action Concept Network

**Acquisition Model**

The acquisition model tightly couples comprehension of utterances with acquisition through metacognitively motivated learning. To comprehend an utterance, the learner exploits all its knowledge to generate a situated interpretation of the utterance and to apply it to the current situation. It may reach different kinds of knowledge failures such as impasses or retrieval failures. In a substate, the learner per-forms a metacognitive analysis. During this analysis, the learner reasons about its current task knowledge and identifies the concept it lacks to generate an appropriate question. It also records how the missing concept relates to its prior knowledge as *context* on the interaction stack, which is useful in assimilating instructions (requirement R4).

### Declarative Knowledge Acquisition

When the learner is given an action command - *"store the blue cylinder,"* it uses its knowledge of English syntax to extract the surface features of the verb and creates a cue – (`<q> ^verb store ^direct-object <o>`). The learner searches its semantic memory for a conceptual graph that matches this cue. If the learner has had no prior experience with the syntactical structure of *store*, this search fails. The learner then deliberately stores this graph in its semantic memory, subgraph (L22,O1), and associates it with a new task operator *op1*, via the mapping node M22 and the object slot O1. The node O1 represents a slot that can be filled up by any object in the environment that satisfies the *direct-object* referring expression (RE) of *store*. Future accesses using this cue allow the learner to access the task operator and the constraints on its instantiation.

If the search of semantic memory is successful, the learner initializes the operator arguments under the constraints afforded by the mapping graph. In the example case, *op1* will be instantiated with an object in the environment that satisfies the referring expression – *the blue cylinder*. This task-operator is then proposed and the learner attempts to execute this operator. The first step in applying a task-operator is generating a *desired state*, which describes the state and the spatial predicates that have to be true in the world for successful completion of a task. To generate the desired state for the current situation, the learner should know what state *storing a blue cylinder* results in. This information is encoded in the tCN as goal predicates. If this is the first time the learner is attempting to execute the action *store*, it does not have the goal concept in its tCN, and therefore it fails to generate the *desired state*.

A summary of why the learner failed is generated and stored on the interaction stack. This summary includes links to the partially built tCN for *store*. The learner queries the expert for the goal of the verb that it is trying to execute. The expert uses the current perceptual and task context to reply, *"The goal is the blue cylinder is in the pantry, and the pantry is closed."* Using its knowledge of grounded prepositions and nouns, the learner extracts spatial and state predicates from this. The predicates are then incorporated in the tCN (accessed using the context summary on the interaction stack) by adding a new subgraph (M22, G22, PR1, PR2, R2, O1, O2).

With the definition of the goal and the constraints afforded by the tCN, the learner can successfully generate an internal representation of the *desired state*. However, it still cannot execute the task and an impasse occurs because the behavior rules are unknown. Therefore, the learner stores the current summary on the interaction stack, and queries the expert for an action that it can take to make progress towards the goal. On a reply – *"open the pantry,"* the learner executes this action in the environment. Using the context on the interaction stack, this action is added to the *problem space* of *store* (subgraph (M22, PS22, A12)). Since, executing this action does not achieve the goal, an impasse occurs again. This interleaving of interaction and execution continues until the *desired state* is achieved in the environment. The learner's episodic memory automatically stores the interactions with the expert, the tasks performed in the environment, and the corresponding state changes.

### Proceduralization

On successfully achieving the *desired state* in the environment, the learner analyzes its experience to explain to itself why this particular sequence of actions achieved the *desired state*. This analysis leads to the creation of rules (through chunking) that implement the instructed behavior. Our approach implements a form of EBL.

During explanation, the learner decouples its internal processing from the external state and retrieves from episodic memory the state it was in when it began the instructed execution of the task. It then attempts to internally simulate the execution of the actions and constituent tasks associated with the new task so that it can use EBL to learn the behavior. First it retrieves the goal definition of the task and instantiates the desired state. It then retrieves the problem-space definition and proposes all associated operators. Chunking compiles this process into rules that propose those operators when the task is being performed. For example, the learner learns that if it is trying to *store* an object, it should consider *moving* it. The learner then uses its memory of the instructions to recursively forward project the problem space actions. If the desired state is achieved, the operator's forward projection is a success and chunking compiles a rule. Each rule tests for the state conditions under which application of an operator resulted in the desired state, and creates a preference for that operator so it is preferred over the other applicable operators. The set of rules complied from this deliberate explanation implement the behavior corresponding to the task.

## Evaluation

Our experiments and demonstrations evaluate the model along the dimensions of behavioral requirements identified earlier. We use the same learner model for all the experiments. The learner's prior knowledge is varied to evaluate different learning behaviors. Prior knowledge states are categorized as: *null,* the learner is not pre-encoded with any domain knowledge beyond the primitive actions; *O*, the learner has prior knowledge of how to recognize objects referred to using noun phrases such as *the red large triangle*; *O+S*, the learner has prior knowledge of object recognition and spatial relationships such as *in*; and *O+S+T*, the learner has declarative and procedural knowledge of the task in addition to the knowledge of object attributes and spatial relationships.

The learner learned and executed three task templates: *move [obj] to [loc]*; *shift [obj] to [loc]*; and *store [obj]*, where *move* is a constituent task of *shift* and *store*. For teaching, task templates were instantiated with an object and a location. If the learner asked any questions, appropriate answers were provided. To acquire and execute these tasks, the learner must comprehend instructions that provide example executions of actions such as *pick up the red large triangle* and goal concepts such as *the goal is the triangle is in the pantry*. This requires the learner to resolve noun phrases to objects in the scene, prepositions to spatial relationships, and verbs to tCNs. The results reported in this paper were obtained from a simulation of the table-top domain that is faithful to the sensory and control systems of the real robot.

### Generality of Task Learning

The results summarized in Table 1 show the knowledge acquired for each task. Column *I* records the number of possible instantiations of a template and *S* records the number of possible initial world states. The acquired knowledge is general (behavior B1) in the following ways.

**Generality of arguments**: Since the learner acquires general tCNs and learns behavior through chunking, the knowledge learned during the single training instance generalizes to other objects and locations in the scene. This was tested by giving randomly instantiated tasks to the learner and verifying that they were correctly executed. Further analysis of the learned rules showed that the learner can execute all instantiations of task after learning from a single example execution with the expert.

**Generality of instruction sequence**: During proceduralization, the learner determines why a sequence of actions and constituent tasks was useful for achieving the desired goal of the task. Superfluous actions in the training sequence, those that were not useful in progress towards the goal, are automatically eliminated. For example, if the training sequence of *move the blue cylinder to the garbage* was *pick up the blue cylinder*, *put the cylinder in the pantry*, *pick up the cylinder*, *put the cylinder in the garbage*, the learner's acquired behavior for *move[obj]*, is *pick-up[obj]*, *put-down[in, obj, loc]* since executing this sequence directly achieves the goal of the task.

**State sensitive task execution**: From a single training sequence, the learner acquires a behavior that applies to any legal initial state (column S) of the world. If the learner is executing *store*, and the *pantry* is *open*, it does not *open* the *pantry* again, but *move*s the relevant object to the *pantry*. Thus the learner acquires causal execution knowledge as opposed to rote memorization of action/task sequences.

Table 1. Tasks acquired by the learner

| Task | I | Goal | Problem Space, Behavior | S |
|---|---|---|---|---|
| move | 16 | in(obj,loc) | pick up (obj), put down (in, obj, loc) | 2 |
| shift | 16 | in(obj,loc) | move (in, obj, loc) | 4 |
| store | 4 | in(obj,pantry) close(pantry) | open (pantry), move (in, obj, pantry), close (pantry) | 4 |

### Changes in Processing

The learner's procedural memory encodes the knowledge for the following cognitive capabilities: interaction management, lexical and referential processing, and learning. To acquire and execute task knowledge from situated interactive instruction, the learner uses these capabilities by selecting associated operators during various stages of processing the instruction. The use of these cognitive capabilities varies with the domain knowledge the learner possesses. Figure 3 shows the distribution of capability use when the learner performs *store the blue cylinder*. In the *null* state, it must not only learn how to perform the task, but it must also acquire knowledge of the spatial-relationship *in*, and of attributes *blue* and *cylinder*. This requires interactions with the expert about these concepts and lexical and comprehension of expert's utterances. When more knowledge is available in the learner ($O+S$, $O+S+T$), it does not need to communicate with the expert to learn these concepts. Therefore, fewer operators belonging to these capabilities are selected. Similarly, object/spatial learning and task-acquisition operators are not selected when the learner possess knowledge of the domain and the task. This demonstrates how processing changes with learning, so that at the end, task-execution dominates. It also shows how the overall time required to perform the task greatly decreases with experience.

These results can be compared to those reported by Anderson (2007) for learning from pre-encoded declarative instructions. With practice in the domain, the instruction interpretation (memory retrieval) steps in ACT-R models dropout. In our model, instruction interpretation corresponds to interaction-management and lexical and referential comprehension. As the model accumulates more knowledge of the task, these capabilities are employed less often.

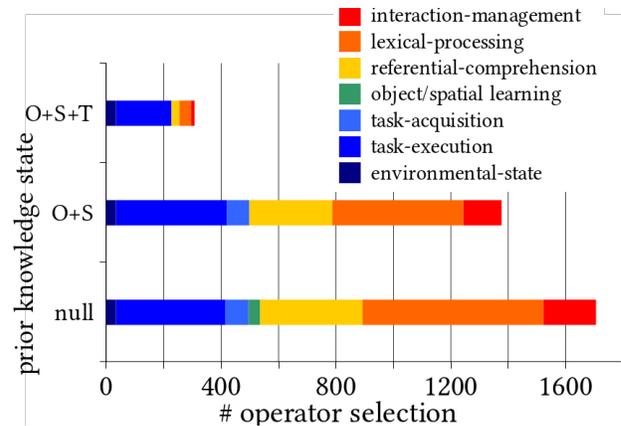

Figure 3. Operator selection during the execution of the task *store* in different initial states of prior knowledge.

### Flexible Instruction

To evaluate if the learner model supports flexible instruction, we report the number of expert-learner utterances required to learn the different actions in each prior knowledge state (Figure 4). Expert-learner utterances are categorized by the knowledge that the learner intends to acquire. For exam-ple, through *object-attribute* utterances, the learner acquires the knowledge of concepts such as *red* and how they map to perceptual symbols.

When the learner begins in *null* state, it initiates several sub-dialogs to learn about objects and spatial relationships. These interactions do not occur if the learner begins in O+S states because it already has knowledge of these concepts. The results show that the learner only requests knowledge when it is missing and therefore, the expert does not need to maintain a perfect mental model of the learner in order to teach it a new concept. The expert can rely on the learner to guide interactions to learning basic concepts if they are unknown to the learner allowing for flexible instruction (behavior B2).

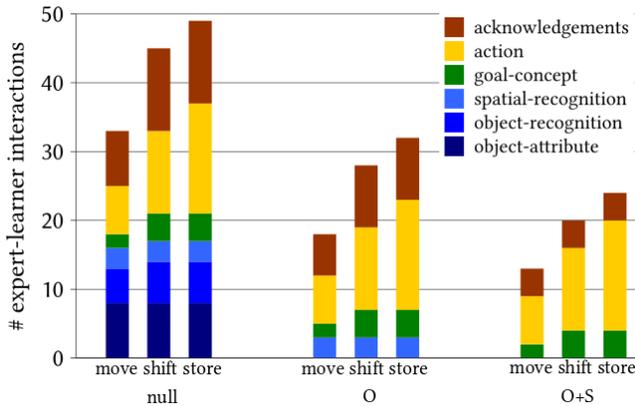

Figure 4. The number of expert-learner interactions when learning tasks in different states of prior knowledge.

**Extendible behavior**

Using the acquired concepts associated with verbs, prepositions, nouns, and adjectives the learner can interactively learn the specifications for problems, such as classic puzzles like 8-Puzzle or Towers of Hanoi, and then solve them (behavior B3). The learner acquires the rules of the problem by asking the expert to specify legal actions in the game, which include what task they are associated with and the preconditions for the action. For example a legal move in 8-Puzzle involves the task *move*, with the conditions that the target location be *empty* and *next-to* the location that the target object is in. The conditions can be either spatial, i.e. "*the block is not in a location*," or describe desired attributes, i.e. "*the block is blue*." The learner also acquires the goal of the problem in a similar manner. The specifications for goals and actions can involve any number of parameters and associated constraints between them, as demonstrated below. The learner requests teaching examples when encountering unknown concepts, while taking advantage of concepts already acquired through other interactions.

A sample of part of the interaction for learning an action for 8-puzzle between Learner *L* and Expert *E* is as follows:

"*L: What in the name of a verb associated with this action?*"
"*E: move*" (initiates teaching if unknown)
"*L: What is a parameter for this action? (or finished if done)*"
"*E: a block*"
"*L: What is a condition for this parameter? (or finished if done)*"
"*E: the block is in 3*" (3 refers to the third parameter)

Using this information, the learner extracts the relevant relationships, indexes potential objects, and determines which actions and goals are present. Even though the expert does not teach full action models of the game mechanics, such as that moving piece `x` from location `y` to `z` in 8-puzzle also causes a `in(x,y)` to be false, the fact that the action knowledge is grounded enables the learner to internally simulate the actions in SVS and search forward for a solution. This enables the learner to solve novel problems where it only receives instruction on the problem definition. Our approach is sufficient to specify a variety of problems that can be described by spatial and visual attribute constraints, or problems with an appropriate isomorphism. To date these include Towers of Hanoi, the Toads and Frogs puzzle, Tic-Tac-Toe, Connect Four, and 8-Puzzle. In future work we will attempt to have the learner learn action models so that more advanced problem solving techniques can be utilized.

## Discussion and Conclusion

Learning from social interactions such as instructions is a distinctly human capability. Through instructions, humans learn a wide range of knowledge, from declarative knowledge in schools to procedural knowledge through apprenticeship with an expert. Prior ACT-R models of learning with instruction (Anderson, 2007) have addressed how procedural knowledge can be acquired through problem solving experience guided by declarative instructions in memory. Such models provided limited answers to how humans learn from social interactions. In this paper, we identified the computational challenges associated with designing models that can learn from mixed-initiative, situated interactions with an expert. We presented a model implemented within the constraints of the Soar cognitive architecture that can learn novel task knowledge through situated instructions. The model acquires diverse knowledge – linguistic, semantic, and procedural by employing different cognitive mechanisms including semantic and episodic memory and chunking.